\documentclass{article}
\usepackage{spconf,amsmath,graphicx,hyperref}
\usepackage{amssymb,mathtools,amsthm}
\usepackage[capitalize]{cleveref}
\usepackage{algorithm}
\usepackage{threeparttable}
\usepackage{booktabs}
\usepackage[noend]{algpseudocode}
\usepackage{setspace}
\usepackage{multirow, makecell}
\usepackage{multicol}
\usepackage{bm}
\crefname{equation}{}{}

\title{Two-Stage Grid Optimization for Group-wise Quantization of LLMs}
\name{Junhan Kim\thanks{$^{\dagger}$Corresponding Author}, Gukryeol Lee, Seungwoo Son, Jeewook Kim, and Yongkweon Jeon$^{\dagger}$}
\address{Samsung Research
\\ \normalsize{junhankim@islab.snu.ac.kr, \{gukryeol.lee, dragwon.jeon\}@samsung.com}}
\begin{document}
%
\maketitle
\begin{abstract}
Group-wise quantization is an effective strategy for mitigating accuracy degradation in low-bit quantization of large language models (LLMs). 
Among existing methods, GPTQ has been widely adopted due to its efficiency; however, it neglects input statistics and inter-group correlations when determining group scales, leading to a mismatch with its goal of minimizing layer-wise reconstruction loss.
In this work, we propose a two-stage optimization framework for group scales that explicitly minimizes the layer-wise reconstruction loss.
In the first stage, performed prior to GPTQ, we initialize each group scale to minimize the group-wise reconstruction loss, thereby incorporating input statistics.
In the second stage, we freeze the integer weights obtained via GPTQ and refine the group scales to minimize the layer-wise reconstruction loss.
To this end, we employ the coordinate descent algorithm and derive a closed-form update rule, which enables efficient refinement without costly numerical optimization.
Notably, our derivation incorporates the quantization errors from preceding layers to prevent error accumulation.
Experimental results demonstrate that our method consistently enhances group-wise quantization, achieving higher accuracy with negligible overhead.
\end{abstract}
\section{Introduction}

To reduce memory usage and accelerate the inference of large language models (LLMs), quantization has received significant attention recently~\cite{frantar2022gptq, ashkboos2024quarot}.
Quantization methods can be broadly classified into two classes: quantization-aware training (QAT)~\cite{jacob2018quantization} and post-training quantization (PTQ)~\cite{nagel2020up}. 
Although QAT achieves high accuracy by incorporating quantization effects during training, it requires extensive data and expensive retraining.
In contrast, PTQ directly quantizes pre-trained models using a small calibration dataset, making it far more practical for billion-scale LLMs~\cite{frantar2022gptq, ashkboos2024quarot, aespa, boa, cushioncache}.

Among PTQ methods, GPTQ has been widely used for its efficiency, quantizing billion-scale LLMs within only a few hours on a single GPU~\cite{frantar2022gptq}. 
In GPTQ, group-wise quantization is often employed to supplement performance in low-bit regimes (e.g., INT2).
This strategy partitions each output channel $\mathbf{w}$ into multiple groups $\mathbf{w}_{1}, \ldots, \mathbf{w}_{n_{g}}$, each of which is assigned a distinct scale $s_{i}$ (see \cref{fig:notation}) to mitigate accuracy loss caused by high intra-channel variance.
A key limitation of GPTQ is that it determines scales based solely on the weight perturbation $\| \Delta \mathbf{w}_{i} \|_{2}^{2}$ of an individual group, completely ignoring input statistics $\mathbf{X}$ and inter-group correlations.
Consequently, the scales are misaligned with GPTQ's objective of minimizing the layer-wise reconstruction loss $\| \Delta \mathbf{w}^{T} \mathbf{X} \|_{2}^{2}=\| \sum_{i} \Delta \mathbf{w}_{i}^{T} \mathbf{X}_{i} \|_{2}^{2}$, leading to suboptimal performance.

In this work, we propose a two-stage optimization framework for group scales that explicitly minimizes the target layer-wise reconstruction loss.
In the first stage, conducted prior to GPTQ, we initialize each group scale $s_{i}$ to minimize the group-wise loss $\| \Delta \mathbf{w}_{i}^{T} \mathbf{X}_{i} \|_{2}^{2}$, thereby accounting for input statistics $\mathbf{X}_{i}$. 
In the second stage, we freeze the integer weights obtained via GPTQ and refine the scales to minimize the total layer-wise loss $\| \sum_{i} \Delta \mathbf{w}_{i}^{T} \mathbf{X}_{i} \|_{2}^{2}$.
To this end, we employ the coordinate descent algorithm, iteratively updating one scale at a time while keeping the others fixed.
We derive a closed-form update rule that enables efficient refinement without the need for costly numerical optimization.
Notably, our derivation incorporates the quantization errors from preceding layers to prevent error accumulation across the network.
Experimental results demonstrate that our method consistently improves accuracy in group-wise quantization with negligible computational overhead.
\vspace{-.2cm}
\section{Background}
\label{sec:background}
\vspace{-.2cm}
\subsection{GPTQ}

GPTQ is a PTQ algorithm extensively adopted for the efficient quantization of LLMs~\cite{frantar2022gptq, ashkboos2024quarot}.
In essence, GPTQ aims to preserve the output of each layer, which is formulated as a layer-wise reconstruction problem:
\begin{align}
    \min_{\mathbf{q} \in \mathcal{Q}}~& \mathbb{E} \hspace{-.7mm} \left [ \left \| \left ( \mathbf{q} - \mathbf{w} \right )^{T} \mathbf{X} \right \|_{2}^{2} \right ]
        &\hspace{-3mm}= \min_{\mathbf{q} \in \mathcal{Q}}~& ( \mathbf{q} - \mathbf{w} )^{T} \mathbf{H} ( \mathbf{q} - \mathbf{w} ), \label{eq:layer-wise reconstruction}
\end{align}
where $\mathbf{q}$ denotes the quantized weights and $\mathbf{H} = \mathbb{E} [\mathbf{X} \mathbf{X}^{T}]$ is the Hessian approximation. 
To solve this, GPTQ first determines the quantization grid (parameterized by scale factors) and then iteratively performs quantization coupled with error compensation.
In each iteration, a single weight is quantized, and the remaining unquantized weights are updated to compensate for the induced error using a Hessian-based update rule~\cite{frantar2022gptq}.
By avoiding time-consuming backpropagation, GPTQ can quantize billion-scale LLMs within a few hours on a single GPU.

\subsection{Group-wise Quantization}

Typically, channel-wise quantization, which assigns a single scale to each output channel, is favored for its efficiency and broad hardware support.
However, at extreme low bit-widths (e.g., INT2), it often incurs severe accuracy degradation because outliers or high intra-channel variance can make scale selection highly susceptible to distribution skew.

\begin{figure}[t]
    \centering
    \centerline{\includegraphics[width=5cm]{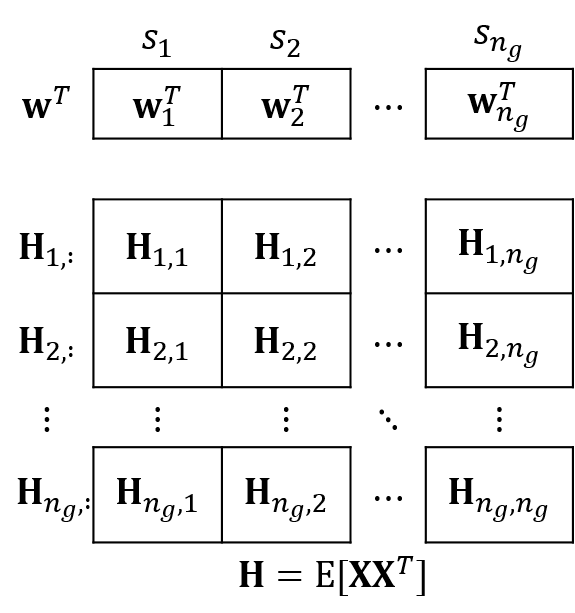}}
    \vspace{-.4cm}
    \caption{Illustration of one output channel in a weight matrix and Hessian for group-wise quantization}
    \label{fig:notation}
    \vspace{-.3cm}
\end{figure}

Group-wise quantization alleviates this by partitioning $\mathbf{w}$ into multiple groups $\mathbf{w}_{1}, \ldots, \mathbf{w}_{n_{g}}$ and assigning a separate scale $s_{i}$ to each group $\mathbf{w}_{i}$ (see \cref{fig:notation}).
By allocating distinct scales, the accuracy degradation from high intra-channel variance can be reduced with only a modest increase in dequantization overhead.
This strategy is now widely adopted in recent PTQ methods~\cite{frantar2022gptq, lin2024awq} and supported by major LLM inference frameworks such as vLLM~\cite{vllm2025} and TensorRT-LLM~\cite{tensorrtllm}.

\subsection{Limitation of GPTQ in Group-wise Quantization}

Before the iterative process, GPTQ determines the quantization grid by optimizing scales to minimize the layer-wise reconstruction loss in~\cref{eq:layer-wise reconstruction}, under the assumption that the nearest quantized value is assigned to each weight~\cite{jeon2023frustratingly}.
For channel-wise quantization, the scale optimization problem is formulated as
\begin{align} \label{eq:optimization problem_channel-wise quantization}
    \begin{split}
    \min_{s > 0}~~~& ( s\mathbf{w}_{int} - \mathbf{w} )^{T} \mathbf{H} ( s\mathbf{w}_{int} - \mathbf{w} ), \\
    \text{s.t. }~~~& \mathbf{w}_{int} = \text{clamp} \left ( \left \lceil \frac{\mathbf{w}}{s} \right \rfloor, 0, 2^{b}-1 \right ),
    \end{split}
\end{align}
where $b$ is the target bit-width.
In practice, $s$ is expressed via a clipping factor $\beta$ with $\beta$ initialized to one,\footnote{$s=\beta \cdot (\max(\mathbf{w}) - \min(\mathbf{w})) / (2^{b} - 1)$} and the optimal $\beta$ is determined through a grid search.
For group-wise quantization, the presence of multiple scales $s_{1}, \cdots, s_{n_{g}}$ transforms the objective in~\cref{eq:optimization problem_channel-wise quantization} into
\begin{align} \label{eq:optimization problem_group-wise quantization}
    \mathcal{L}(\mathbf{s})
        &= \sum_{i, j=1}^{n_{g}} ( s_{i}\mathbf{w}_{int, i} - \mathbf{w}_{i} )^{T} \mathbf{H}_{i, j} ( s_{j}\mathbf{w}_{int, j} - \mathbf{w}_{j} ),
\end{align}
where $\mathbf{s} = [s_{1}, \ldots, s_{n_{g}}]^{T}$, $\mathbf{w}_{int,i}$ denotes the integer weights in the $i$-th group, and $\mathbf{H}_{i,j} = \mathbb{E} [ \mathbf{X}_{i} \mathbf{X}_{j}^{T} ]$ represents the cross-correlation between the inputs to groups $i$ and $j$ (see \cref{fig:notation}).

Unlike the channel-wise case, a joint grid search over all scales is computationally prohibitive due to its exponential complexity, $\mathcal{O}(M^{n_{g}})$, where $M$ is the number of candidates for each scale.
To simplify this, GPTQ assumes $\mathbf{H} = \mathbf{I}$ (i.e., $\mathbf{H}_{i, i} = \mathbf{I}$ and $\mathbf{H}_{i,j} = \mathbf{0}$ for $i \neq j$), which reduces~\cref{eq:optimization problem_group-wise quantization} to a sum of independent group-wise terms $\| s_{i}\mathbf{w}_{int, i} - \mathbf{w}_{i} \|_{2}^{2}$.
While this allows for parallel optimization, it fundamentally ignores input statistics $\mathbf{X}$ and inter-group correlations, leading to suboptimal scale selection.
This motivates our method, which explicitly optimizes the scales to minimize the target layer-wise reconstruction loss in~\cref{eq:optimization problem_group-wise quantization}.
\section{Proposed Method}
\label{sec:method}

\subsection{Overview: Two-stage Optimization for Group Scales}

Our method augments GPTQ with two-stage scale refinement to address its suboptimal grid selection.

\textbf{(Stage 1. Input-aware scale initialization)}
This stage, conducted prior to GPTQ's iterative process, aims to initialize each group scale $s_{i}$ by incorporating the corresponding input statistics $\mathbf{X}_{i}$.
Specifically, we optimize $s_{i}$ to minimize a local group-wise reconstruction loss $\| \Delta \mathbf{w}_{i}^{T} \mathbf{X}_{i} \|_{2}^{2}$, formulated as
\begin{align}
    \begin{split}
    \min_{s_{i} > 0}~~~&( s_{i}\mathbf{w}_{int, i} - \mathbf{w}_{i} )^{T} 
    \mathbf{H}_{i, i} 
    ( s_{i}\mathbf{w}_{int, i} - \mathbf{w}_{i} ), \\
    \text{s.t. }~~~&\mathbf{w}_{int, i} = \text{clamp} \left ( \left \lceil \frac{\mathbf{w}_{i}}{s_{i}} \right \rfloor, 0, 2^{b}-1 \right ).
    \end{split}
\end{align}
This problem is separable across groups and can therefore be solved efficiently in parallel via independent grid searches.
In addition, this step requires no additional memory or computation for $\mathbf{H}_{i,i}$, as it can be directly extracted from the pre-computed Hessian $\mathbf{H}$ (see \cref{fig:notation}), which is already available within the standard GPTQ pipeline.

\textbf{(Stage 2. Scale refinement)}
After the iterative process of GPTQ, we \emph{freeze} the integer weights $\mathbf{w}_{int}$ and then refine the group scales $\mathbf{s}$ to minimize the layer-wise reconstruction loss in~\cref{eq:optimization problem_group-wise quantization}.
By doing so, inter-group correlations, which are not considered in the first stage, can be accounted for, while efficiency is maintained because only the scales (rather than the integer weights $\mathbf{w}_{int}$) are updated.
To accomplish this, we employ the \emph{coordinate descent} (CD) algorithm, iteratively updating one scale at a time while keeping the others fixed.
Each update step aims to minimize the layer-wise reconstruction loss in~\cref{eq:optimization problem_group-wise quantization}, thereby yielding scales that are most consistent with the true optimization goal.
In the following subsections, we present a closed-form update rule, making the proposed method mathematically grounded and computationally efficient.

\subsection{Scale Refinement for the First Layer}

Given the assigned integers $\mathbf{w}_{int}$, we keep them fixed and apply CD to refine the group scales $\mathbf{s}$.
Since the objective is quadratic in each $s_{i}$, the optimal $s_{i}^{*}$ in each CD step can be obtained by setting $\partial \mathcal{L}/\partial s_{i}=0$.
Noting that
\begin{align*}
    \frac{\partial \mathcal{L}}{\partial s_{i}}
        &= (2\mathbf{w}_{int, i}^{T} \mathbf{H}_{i, i} \mathbf{w}_{int, i})s_{i} + 2\sum_{j \neq i} (\mathbf{w}_{int, i}^{T} \mathbf{H}_{i, j} \mathbf{w}_{int, j}) s_{j} \nonumber \\
        &~~~- 2 \sum_{j=1}^{n_{g}} (\mathbf{w}_{int, i}^{T} \mathbf{H}_{i, j} \mathbf{w}_{j}),
\end{align*}
the closed-form update rule for $s_{i}$ is derived as
\begin{align}
    s_{i}^{*} 
        &= \frac{\sum_{j=1}^{n_{g}} \mathbf{w}_{int, i}^{T} \mathbf{H}_{i, j} \mathbf{w}_{j} - \sum_{j \neq i} (\mathbf{w}_{int,i}^{T} \mathbf{H}_{i, j} \mathbf{w}_{int, j}) s_{j}}{\mathbf{w}_{int, i}^{T} \mathbf{H}_{i, i} \mathbf{w}_{int, i}} \nonumber \\
        &= s_{i} + \frac{\mathbf{w}_{int, i}^{T} \sum_{j=1}^{n_{g}} \mathbf{H}_{i, j} (\mathbf{w}_{j} - s_{j} \mathbf{w}_{int, j})}{\mathbf{w}_{int, i}^{T} \mathbf{H}_{i, i} \mathbf{w}_{int, i}}. \nonumber
\end{align}
Let $\mathbf{H}_{i, :} = [ \mathbf{H}_{i, 1}, \ \ldots, \ \mathbf{H}_{i, n_{g}}]$, and let $\mathbf{q}_{j} = s_{j} \mathbf{w}_{int, j}$ be the current quantized weights, then the update rule simplifies to
\begin{align}
    s_{i}^{*}
        &= s_{i} + \frac{\mathbf{w}_{int, i}^{T} \mathbf{H}_{i, :} (\mathbf{w} - \mathbf{q})}{\mathbf{w}_{int, i}^{T} \mathbf{H}_{i, i} \mathbf{w}_{int, i}}. \label{eq:update formula_group_first}
\end{align}
Notably, all required terms $\mathbf{H}_{i, i}$ and $\mathbf{H}_{i, :}$ are obtainable directly from the pre-computed Hessian $\mathbf{H}$ (see \cref{fig:notation}), requiring no additional memory or computation.
For channel-wise quantization where $n_{g}=1$,~\cref{eq:update formula_group_first} reduces to
\begin{align}
    s^{*}
        &= s + \frac{\mathbf{w}_{int}^{T} \mathbf{H} (\mathbf{w} - \mathbf{q})}{\mathbf{w}_{int}^{T} \mathbf{H} \mathbf{w}_{int}}
        = \frac{\mathbf{w}_{int}^{T} \mathbf{H} \mathbf{w}}{\mathbf{w}_{int}^{T} \mathbf{H} \mathbf{w}_{int}}, \label{eq:update formula_channel}
\end{align}
which coincides with the result in~\cite{comq}.

\begin{table}[!t]
\begin{threeparttable}
\begin{algorithm}[H]
\begin{spacing}{1.1}
\caption{Group-Scale Refinement}
\label{algo:cd}
\renewcommand\algorithmicrequire{\textbf{Input}:}
\renewcommand\algorithmicensure{\textbf{Output}:}
\begin{algorithmic}[1]
\Require FP weights $\mathbf{w}$, integer weights $\mathbf{w}_{int}$, number $n_{g}$ of groups, group size $g$, initial scales $\mathbf{s}$, Hessian $\mathbf{H}=\mathbb{E}[\mathbf{X}\mathbf{X}^{T}]$, deviation correlation $\mathbf{R}=\mathbb{E}[\Delta\mathbf{X}\mathbf{X}^{T}]$
\Ensure Quantized weights $\mathbf{q}$, optimized scales $\mathbf{s}$

\State Initialize quantized weights: $\mathbf{q} \leftarrow \mathbf{s} \odot_{g} \mathbf{w}_{int}$ 
\For{$i=0, \cdots, n_{g}-1$}
    \State Extract $i$-th group: $\mathbf{w}_{int,i} \leftarrow \mathbf{w}_{int}[ig:(i+1)g]$
    \State Extract Hessian blocks:
    \Statex \hspace{6mm} $\mathbf{H}_{i,:} \leftarrow \mathbf{H}[ig:(i+1)g, :]$
    \Statex \hspace{6mm} $\mathbf{H}_{i,i} \leftarrow \mathbf{H}[ig:(i+1)g, ig:(i+1)g]$
    \State Extract deviation term: $\mathbf{R}_{i} \leftarrow \mathbf{R}[:, ig:(i+1)g]$
    \State \textbf{Update scale for group $i$:}
    \vspace{-2.5mm}
        \begin{equation*}
           s_i \leftarrow 
           s_{i} + \frac{\mathbf{w}_{int, i}^{T} \mathbf{H}_{i, :} (\mathbf{w} -\mathbf{q})- \mathbf{w}^{T} \mathbf{R}_{i} \mathbf{w}_{int, i} }{\mathbf{w}_{int, i}^{T} \mathbf{H}_{i, i} \mathbf{w}_{int, i}}. 
        \end{equation*}
    \vspace{-3.5mm}
    \State Update quantized weights: $\mathbf{q} \leftarrow \mathbf{s} \odot_{g} \mathbf{w}_{int}$ 
\EndFor
\end{algorithmic}
\end{spacing}
\end{algorithm}
\vspace{-5mm}
\begin{tablenotes}
    \item[] $\odot_g$ means that each scale $s_i$ multiplies the corresponding group $\mathbf{w}_{int,i}$.
\end{tablenotes}
\end{threeparttable}

\vspace{-.3cm}
\end{table}

\subsection{Quantization Error-aware Scale Refinement for Subsequent Layers}

In all layers following the first, the input $\mathbf{X}$ differs from the full-precision (FP) input $\widetilde{\mathbf{X}}$ due to the quantization of preceding layers.
The input deviation $\Delta \mathbf{X} = \mathbf{X} - \widetilde{\mathbf{X}}$ can be incorporated into the layer-wise reconstruction error as follows:
\begin{align}
    \mathcal{L}(\mathbf{s})
        &\hspace{-.5mm}=\hspace{-.5mm} \mathbb{E} [\| \mathbf{q}^{T} \mathbf{X} - \mathbf{w}^{T} \widetilde{\mathbf{X}} \|_{2}^{2}] 
        = \mathbb{E} \hspace{-1mm} \left [ \| (\mathbf{q} - \mathbf{w})^{T} \mathbf{X} + \mathbf{w}^{T} \hspace{-.7mm} \Delta \hspace{-.57mm} \mathbf{X}\|_{2}^{2} \right ] \nonumber \\
        &\hspace{-.5mm}=\hspace{-.5mm} (\mathbf{q} - \mathbf{w})^{T} \mathbf{H} (\mathbf{q} - \mathbf{w}) + 2 \mathbf{w}^{T} \mathbf{R} (\mathbf{q} - \mathbf{w}) + c, \label{eq:refined objective_group-wise_later}
\end{align}
where $\mathbf{R} = \mathbb{E} [\Delta \mathbf{X} \mathbf{X}^{T}]$ and $c$ is a constant with respect to $\mathbf{s}$.
Compared to the loss for the first layer (see~\cref{eq:layer-wise reconstruction}), the new loss involves an additional term $\mathbf{w}^{T} \mathbf{R} (\mathbf{q} - \mathbf{w})$, which explicitly captures the errors introduced by the input deviation $\Delta \mathbf{X}$.
Noting that the derivative of this new term is
\begin{align}
    \frac{\partial\mathbf{w}^{T} \mathbf{R} (\mathbf{q} - \mathbf{w})}{\partial s_{i}}
        &= \mathbf{w}^{T} \mathbf{R}_{i} \mathbf{w}_{int, i},
\end{align}
where $\mathbf{R}_{j} = \mathbb{E} [\Delta \mathbf{X} \mathbf{X}_{j}^{T}]$, the update rule for subsequent layers is refined to
\begin{align}
    s_{i}^{*}
        &= s_{i} + \frac{\mathbf{w}_{int, i}^{T} \mathbf{H}_{i, :} (\mathbf{w} -\mathbf{q})- \mathbf{w}^{T} \mathbf{R}_{i} \mathbf{w}_{int, i} }{\mathbf{w}_{int, i}^{T} \mathbf{H}_{i, i} \mathbf{w}_{int, i}}. \label{eq:update formula_group_later}
\end{align}
\cref{algo:cd} summarizes the pseudocode for the proposed CD-based scale refinement stage.
\section{Experiments}

\begin{table}[t]
    \renewcommand{\arraystretch}{1.0}
    \scriptsize
    \centering
    \caption{Group-wise quantization on Llama (group size=64)}
    \begin{threeparttable}
    \begin{tabular}{c c c c c c c c c c c c c}
    \toprule
    \multirowcell{1}{Model} & \multirowcell{1}{Precision} & \multirowcell{1}{Method} & \multirowcell{1}{Wiki2 ($\downarrow$)} & \multirowcell{1}{C4 ($\downarrow$)} & \multirowcell{1}{0-shot ($\uparrow$)} \\
    \toprule
    \multirowcell{6}{Llama3.2-1B\\-Instruct}
    & FP & baseline & 13.16 & 21.31 & 56.82 \\
    \cmidrule{2-6}
    & \multirowcell{2}{INT2}
    & GPTQ & 214.7 & 429.4 & 32.00 \\
    & & \textbf{ours} & \textbf{63.31} & \textbf{149.5} & \textbf{33.62} \\
    \cmidrule{2-6}
    & \multirowcell{2}{INT3}
    & GPTQ & 18.15 & 30.73 & 50.61 \\
    & & \textbf{ours} & \textbf{16.30} & \textbf{26.83} & \textbf{53.19} \\
    \midrule
    \multirowcell{6}{Llama3.2-3B\\-Instruct}
    & FP & baseline & 11.05 & 16.49 & 63.01 \\
    \cmidrule{2-6}
    & \multirowcell{2}{INT2}
    & GPTQ & 73.70 & 130.2 & 38.55 \\
    & & \textbf{ours} & \textbf{29.33} & \textbf{80.76} & \textbf{42.73} \\
    \cmidrule{2-6}
    & \multirowcell{2}{INT3}
    & GPTQ & 13.43 & 19.69 & 59.68 \\
    & & \textbf{ours} & \textbf{12.66} & \textbf{19.58} & \textbf{61.33} \\
    \midrule
    \multirowcell{6}{Llama3-8B}
    & FP & baseline & 6.139 & 9.444 & 70.34 \\
    \cmidrule{2-6}
    & \multirowcell{2}{INT2}
    & GPTQ & 15.07 & 209.7 & 43.60 \\
    & & \textbf{ours} & \textbf{13.84} & \textbf{33.14} & \textbf{49.56} \\
    \cmidrule{2-6}
    & \multirowcell{2}{INT3}
    & GPTQ & 7.202 & 12.53 & 66.88 \\
    & & \textbf{ours} & \textbf{7.131} & \textbf{12.14} & \textbf{68.74} \\
    \midrule
    \multirowcell{6}{Llama2-7B}
    & FP16 & baseline & 5.473 & 7.266 & 67.28 \\
    \cmidrule{2-6}
    & \multirowcell{2}{INT2}
    & GPTQ & 13.67 & 40.27 & 50.35 \\
    & & \textbf{ours} & \textbf{8.274} & \textbf{13.95} & \textbf{54.95} \\
    \cmidrule{2-6}
    & \multirowcell{2}{INT3}
    & GPTQ & 6.171 & 8.333 & 63.95 \\
    & & \textbf{ours} & \textbf{5.822} & \textbf{8.097} & \textbf{65.14} \\
    \midrule
    \multirowcell{6}{Llama2-13B}
    & FP & baseline & 4.885 & 6.730 & 69.83 \\
    \cmidrule{2-6}
    & \multirowcell{2}{INT2}
    & GPTQ & 7.248 & 12.35 & 57.08 \\
    & & \textbf{ours} & \textbf{6.962} & \textbf{11.88} & \textbf{59.91} \\
    \cmidrule{2-6}
    & \multirowcell{2}{INT3}
    & GPTQ & 5.159 & \textbf{7.320} & 68.15 \\
    & & \textbf{ours} & \textbf{5.155} & 7.323 & \textbf{68.65} \\
    \bottomrule
    \end{tabular}
    \end{threeparttable}
    \label{tab:comparison_with_gptq_64}

    \vspace{-.3cm}

\end{table}

We evaluate our method using the Llama2 and Llama3 models~\cite{touvron2023Llama2}.
Following~\cite{frantar2022gptq}, we perform weight-only quantization while keeping activations in FP, which effectively accelerates LLM inference by reducing memory movement.
As calibration data, we sampled 128 random sequences of length 2048 from WikiText-2 (Wiki2)~\cite{wiki}.
Performance is measured via perplexity (PPL) on the Wiki2 and C4~\cite{c4} test splits and average accuracy on several zero-shot commonsense reasoning tasks.\footnote{ARC-challenge and ARC-easy~\cite{allenai:arc}, BoolQ~\cite{clark2019boolq}, OpenbookQA~\cite{mihaylov2018openbookqa}, LAMBADA~\cite{paperno2016lambada}, PIQA~\cite{bisk2020piqa}, HellaSwag~\cite{zellers2019hellaswag}, and WinoGrande~\cite{sakaguchi2021winogrande}}
All experiments were conducted on a single NVIDIA H100 GPU (80~GB).

\subsection{Comparison with GPTQ}

Tables~\ref{tab:comparison_with_gptq_64} and~\ref{tab:comparison_with_gptq_32} summarize the PPL and zero-shot accuracy for GPTQ and the proposed method.
The group size, the number of consecutive weights sharing a scale factor, is set to 64 for \cref{tab:comparison_with_gptq_64} and 32 for \cref{tab:comparison_with_gptq_32}.
Overall, performance improves as group size decreases due to the increased number of scale factors.
In both cases, the proposed method consistently outperforms GPTQ.
For example, under 2-bit quantization, we achieve 6\%p and 4\%p improvements over GPTQ on Llama3-8B and Llama3.2-3B-Instruct/Llama2-7B, respectively.
For small-scale LLMs like Llama3.2-1B-Instruct, the C4 PPL improves significantly ($429.4\rightarrow149.5$).
Notably, at 3-bit, our method nearly preserves the original FP performance, incurring only a 2\%p accuracy drop.
This superiority stems from our explicit minimization of the layer-wise reconstruction loss, which accounts for input statistics and inter-group correlations.
In contrast, GPTQ determines each group scale independently, thereby ignoring these critical factors.

\begin{table}[t]
    \renewcommand{\arraystretch}{1.0}
    \scriptsize
    \centering
    \caption{Group-wise quantization on Llama3 (group size=32)}
    \begin{threeparttable}
    \begin{tabular}{c c c c c c c c c c c c c}
    \toprule
    \multirowcell{1}{Model} & \multirowcell{1}{Precision} & \multirowcell{1}{Method} & \multirowcell{1}{Wiki2 ($\downarrow$)} & \multirowcell{1}{C4 ($\downarrow$)} & \multirowcell{1}{0-shot ($\uparrow$)} \\
    \toprule
    \multirowcell{6}{Llama3.2-1B\\-Instruct}
    & FP & baseline & 13.16 & 21.31 & 56.82 \\
    \cmidrule{2-6}
    & \multirowcell{2}{INT2}
    & GPTQ & 113.1 & 205.6 & 34.14 \\
    & & \textbf{ours} & \textbf{41.11} & \textbf{87.59} & \textbf{37.21} \\
    \cmidrule{2-6}
    & \multirowcell{2}{INT3}
    & GPTQ & 17.40 & 28.23 & 52.08 \\
    & & \textbf{ours} & \textbf{15.45} & \textbf{25.33} & \textbf{53.91} \\
    \midrule
    \multirowcell{6}{Llama3.2-3B\\-Instruct}
    & FP & baseline & 11.05 & 16.49 & 63.01 \\
    \cmidrule{2-6}
    & \multirowcell{2}{INT2}
    & GPTQ & 51.92 & 70.63 & 41.45 \\
    & & \textbf{ours} & \textbf{22.99} & \textbf{50.50} & \textbf{46.57} \\
    \cmidrule{2-6}
    & \multirowcell{2}{INT3}
    & GPTQ & 14.55 & 19.70 & 59.34 \\
    & & \textbf{ours} & \textbf{12.29} & \textbf{18.94} & \textbf{61.89} \\
    \midrule
    \multirowcell{6}{Llama3-8B}
    & FP & baseline & 6.139 & 9.444 & 70.34 \\
    \cmidrule{2-6}
    & \multirowcell{2}{INT2}
    & GPTQ & 13.22 & 35.78 & 47.09 \\
    & & \textbf{ours} & \textbf{11.52} & \textbf{24.67} & \textbf{53.65} \\
    \cmidrule{2-6}
    & \multirowcell{2}{INT3}
    & GPTQ & 7.417 & 12.33 & 65.49 \\
    & & \textbf{ours} & \textbf{6.920} & \textbf{11.59} & \textbf{69.29} \\
    \bottomrule
    \end{tabular}
    \end{threeparttable}
    \label{tab:comparison_with_gptq_32}

    \vspace{-.3cm}
    
\end{table}

\begin{table}[t]
    \renewcommand{\arraystretch}{1.0}
    \footnotesize
    \centering
    \caption{Ablation study of each stage on 2-bit group-wise quantization of Llama3.2-1B-Instruct (group size=64)}
    \begin{threeparttable}
    \begin{tabular}{c c c c c c c c c c c c c}
    \toprule
    \multirowcell{1}{Method} & \multirowcell{1}{Stage 1} & \multirowcell{1}{Stage 2} & \multirowcell{1}{Wiki2 ($\downarrow$)} & \multirowcell{1}{C4 ($\downarrow$)} & \multirowcell{1}{Time (min)} \\
    \toprule
    GPTQ & & & 214.7 & 429.4 & 5.85 \\
    \midrule
    \multirowcell{3}{\textbf{ours}} 
    & \checkmark & & 130.6 & 194.0 & 6.82 \\
    & & \checkmark & 86.15 & 215.9 & 6.30 \\
    & \checkmark & \checkmark & 63.31 & 149.5 & 7.53 \\
    \bottomrule
    \end{tabular}
    \end{threeparttable}
    \label{tab:ablation}

    \vspace{-.3cm}
    
\end{table}

\subsection{Ablation Study}

The ablation results in~\cref{tab:ablation} highlight the effectiveness of each stage in the proposed scale optimization method. 
By initializing group scales based on input statistics (stage 1), we achieve substantial PPL improvements on both Wiki2 (214.7 → 130.6) and C4 (429.4 → 194.0), with only a minor increase in runtime. 
The scale refinement stage (stage 2) also yields significant gains (Wiki2: 214.7 → 86.15, C4: 429.4 → 215.9) with minimal overhead. 
Importantly, combining both stages delivers the best performance while maintaining runtime within a reasonable range. 
These results clearly demonstrate that the two stages are complementary: Stage 1 incorporates input statistics for more reliable initialization, while Stage 2 explicitly accounts for inter-group correlations, jointly enabling the most accurate quantization.
\section{Conclusion}

In this paper, we presented a two-stage optimization framework for group scales to enhance the group-wise quantization performance of GPTQ.
In the first stage, group scales are initialized to minimize the group-wise reconstruction loss, effectively incorporating input statistics.
In the second stage, conducted after GPTQ's iterative process, we refine the scales to align with the layer-wise reconstruction objective.
To this end, we employed the CD algorithm and derived a closed-form update rule that accounts for quantization errors from preceding layers.
Our experimental results demonstrated that the proposed method consistently improves quantization accuracy with negligible computational overhead.

\vfill
\pagebreak 

\bibliographystyle{IEEEbib}
\bibliography{strings,refs}

@inproceedings{jacob2018quantization,
  title={Quantization and training of neural networks for efficient integer-arithmetic-only inference},
  author={B.~Jacob and S.~Kligys and B.~Chen and M.~Zhu and M.~Tang and A.~Howard and H.~Adam and D.~Kalenichenko},
  booktitle={Proceedings of the IEEE conference on computer vision and pattern recognition},
  pages={2704--2713},
  year={2018}
}

@inproceedings{aespa,
 author = {J.~Kim and C.~Lee and E.~Cho and K.~Park and H.~Kim and J.~Kim and Y.~Jeon},
 booktitle = {Advances in Neural Information Processing Systems (NeurIPS)},
 pages = {94292--94326},
 title = {Towards Next-Level Post-Training Quantization of Hyper-Scale Transformers},
 volume = {37},
 year = {2024}
}

@inproceedings{boa,
    title={{BoA}: Attention-aware Post-training Quantization without Backpropagation},
    author={J.~Kim and H.~Kim and E.~Cho and C.~Lee and J.~Kim and Y.~Jeon},
    booktitle={Forty-second International Conference on Machine Learning (ICML)},
    year={2025},
}

@inproceedings{cushioncache,
  title={Prefixing Attention Sinks can Mitigate Activation Outliers for Large Language Model Quantization},
  author={S.~Son and W.~Park and W.~Han and K.~Kim and J.~Lee},
  booktitle={Proceedings of the 2024 Conference on Empirical Methods in Natural Language Processing},
  pages={2242--2252},
  year={2024}
}

@inproceedings{nagel2020up,
  title={Up or down? {A}daptive rounding for post-training quantization},
  author={M.~Nagel and R.A.~Amjad and M.~Van Baalen and C.~Louizos and T.~Blankevoort},
  booktitle={International Conference on Machine Learning (ICML)},
  pages={7197--7206},
  year={2020}
}

@article{comq,
  title={{COMQ}: A Backpropagation-Free Algorithm for Post-Training Quantization},
  author={A.~Zhang and Z.~Yang and N.~Wang and Y.~Qi and J.~Xin and X.~Li and P.~Yin},
  journal={arXiv:2403.07134},
  year={2024}
}

@article{ashkboos2024quarot,
  title={{QuaRot}: Outlier-free 4-bit inference in rotated {LLMs}},
  author={S.~Ashkboos and A.~Mohtashami and M.L.~Croci and B.~Li and P.~Cameron and M.~Jaggi and D.~Alistarh and T.~Hoefler and J.~Hensman},
  journal={arXiv:2404.00456},
  year={2024}
}

@inproceedings{frantar2022gptq,
  title={{GPTQ}: Accurate post-training quantization for generative pre-trained Transformers},
  author={E.~Frantar and S.~Ashkboos and T.~Hoefler and D.~Alistarh},
  booktitle = {International Conference on Learning Representations},
  year      = {2023}
}

@inproceedings{lin2024awq,
  title={{AWQ}: Activation-aware weight quantization for on-device {LLM} compression and acceleration},
  author={J.~Lin and J.~Tang and H.~Tang and S.~Yang and W.M.~Chen and W.C.~Wang and G.~Xiao and X.~Dang and C.~Gan and S.~Han},
  booktitle={Proceedings of machine learning and systems},
  year={2024}
}

@article{touvron2023llama2,
  title={Llama 2: Open foundation and fine-tuned chat models},
  author={H.~Touvron and L.~Martin and K.~Stone and P.~Albert and A.~Almahairi and Y.~Babaei and N.~Bashlykov and S.~Batra and P.~Bhargava and S.~Bhosale and others},
  journal={arXiv:2307.09288},
  year={2023}
}

@article{wiki,
  title={Pointer sentinel mixture models},
  author={S.~Merity and C.~Xiong and J.~Bradbury and R.~Socher},
  journal={arXiv:1609.07843},
  year={2016}
}

@article{c4,
  title={Exploring the limits of transfer learning with a unified text-to-text transformer},
  author={C.~Raffel and N.~Shazeer and A.~Roberts and K.~Lee and S.~Narang and M.~Matena and Y.~Zhou and W.~Li and P.J.~Liu},
  journal={Journal of Machine Learning Research},
  volume={21},
  number={1},
  pages={5485--5551},
  year={2020},
  publisher={JMLRORG}
}

@article{allenai:arc,
      author    = {P.~Clark and I.~Cowhey and O.~Etzioni and T.~Khot and A.~Sabharwal and C.~Schoenick and O.~Tafjord},
      title     = {Think you have Solved Question Answering? {T}ry {ARC}, the {AI2} Reasoning Challenge},
      journal   = {arXiv:1803.05457v1},
      year      = {2018},
}

@article{clark2019boolq,
  title={{BoolQ}: Exploring the surprising difficulty of natural yes/no questions},
  author={C.~Clark and K.~Lee and M.W.~Chang and T.~Kwiatkowski and M.~Collins and K.~Toutanova},
  journal={arXiv:1905.10044},
  year={2019}
}

@inproceedings{zellers2019hellaswag,
  title={{HellaSwag}: Can a Machine Really Finish Your Sentence?},
  author={R.~Zellers and A.~Holtzman and Y.~Bisk and A.~Farhadi and Y.~Choi},
  booktitle={Proceedings of the 57th Annual Meeting of the Association for Computational Linguistics},
  pages={4791--4800},
  year={2019}
}

@article{mihaylov2018openbookqa,
  title={Can a suit of armor conduct electricity? a new dataset for open book question answering},
  author={T.~Mihaylov and P.~Clark and T.~Khot and A.~Sabharwal},
  journal={arXiv:1809.02789},
  year={2018}
}

@article{sakaguchi2021winogrande,
  title={{WinoGrande}: An adversarial winograd schema challenge at scale},
  author={K.~Sakaguchi and R.L.~Bras and C.~Bhagavatula and Y.~Choi},
  journal={Communications of the ACM},
  volume={64},
  number={9},
  pages={99--106},
  year={2021},
  publisher={ACM New York, NY, USA}
}

@article{paperno2016lambada,
  title={The {LAMBADA} dataset: Word prediction requiring a broad discourse context},
  author={D.~Paperno and G.~Kruszewski and A.~Lazaridou and Q.N.~Pham and R.~Bernardi and S.~Pezzelle and M.~Baroni and G.~Boleda and R.~Fern{\'a}ndez},
  journal={arXiv:1606.06031},
  year={2016}
}

@inproceedings{bisk2020piqa,
  title={{PIQA}: Reasoning about physical commonsense in natural language},
  author={Y.~Bisk and R.~Zellers and J.~Gao and Y.~Choi and others},
  booktitle={Proceedings of the AAAI conference on artificial intelligence},
  volume={34},
  number={05},
  pages={7432--7439},
  year={2020}
}

@misc{vllm2025,
  title   = {{vLLM}: Easy, Fast, and Cheap {LLM} Serving},
  author  = {{vLLM Team}},
  howpublished = {\url{https://docs.vllm.ai}},
  year    = {2025},
  note    = {Accessed: Sept. 2025}
}

@misc{tensorrtllm,
  title   = {{TensorRT-LLM}: High-performance Inference for Large Language Models},
  author  = {{NVIDIA}},
  howpublished = {\url{https://docs.nvidia.com/deeplearning/tensorrt-llm}},
  year    = {2025},
  note    = {Accessed: Sept. 2025}
}

@inproceedings{jeon2023frustratingly,
  title={A frustratingly easy post-training quantization scheme for {LLMs}},
  author={Y.~Jeon and C.~Lee and K.~Park and H.~Kim},
  booktitle={Proceedings of the 2023 Conference on Empirical Methods in Natural Language Processing},
  pages={14446--14461},
  year={2023}
}

\end{document}